\pdfoutput=1
\documentclass[11pt]{article}

\usepackage{acl}

\usepackage{times}
\usepackage{latexsym}
\usepackage{graphicx}
\usepackage{booktabs}
\usepackage{tcolorbox}
\usepackage{amsmath}

\usepackage[T1]{fontenc}
\usepackage[utf8]{inputenc}

\usepackage{microtype}

\usepackage{colortbl}
\usepackage{xcolor}
\title{\dataset: Probing language models for in-context word acquisition}

\author{Julian Martin Eisenschlos$^1$, Jeremy R. Cole$^1$, Fangyu Liu$^2$, William W. Cohen$^1$ \\
$^1$ Google Research \ \ \ \ \  $^2$ University of Cambridge \\
\texttt{\{eisenjulian,jrcole,wcohen\}@google.com \ \ fl399@cam.ac.uk}
 }
\definecolor{darkgreen}{HTML}{008f00}
\definecolor{UMDred}{HTML}{ed1c24}
\newcommand{\palm}{\textsc{PaLM}}
\newcommand{\gptt}{\textsc{GPT-3}}
\newcommand{\dataset}{\textsc{WinoDict}}
\newcommand{\winograd}{\textsc{Winograd}}
\newcommand{\winogrande}{\textsc{WinoGrande}}

\newcommand{\err}[1]{\textsubscript{~$\pm$#1}}

\begin{document}
\maketitle
\begin{abstract}
We introduce a new in-context learning paradigm to measure Large Language Models' (LLMs) ability to learn novel words during inference. In particular, we rewrite Winograd-style co-reference resolution problems by replacing the key concept word with a synthetic but plausible word that the model must understand to complete the task. Solving this task requires the model to make use of the dictionary definition of the new word given in the prompt. This benchmark addresses word acquisition, one important aspect of the diachronic degradation known to afflict LLMs. As LLMs are frozen in time at the moment they are trained, they are normally unable to reflect the way language changes over time. We show that the accuracy of LLMs compared to the original Winograd tasks decreases radically in our benchmark, thus identifying a limitation of current models and providing a benchmark to measure future improvements in LLMs ability to do in-context learning.    

% By the nature of the cost and time required to train Large Language Models (LLMs), the embedded knowledge within is usually frozen at the moment their training data is collected. 
% %
% As a result, LLMs have been shown to suffer from diachronic degradation.
% %
% The in-context learning paradigm can provide a workaround for this limitation by supplying relevant information at inference time. 
% %
% We introduce a new benchmark to evaluate LLMs for one particular but critical aspect of diachronic change: word acquisition.
% %
% To that end, we rewrite Winograd-style co-reference resolution problems by replacing a word for a new synthetic but plausible English word.
% %
% The meaning of the word is given to the model in the prompt via a dictionary definition.
% %
% We show that the accuracy of LLMs compared to the original Winograd tasks decreases radically in our benchmark and we believe this serves as a measure of progress for future models.

\end{abstract}

\section{Introduction}

Large Language Models (LLMs) such as \gptt~\citep{brown2020language} and \palm~\citep{chowdhery2022palm} can only learn from information that is in their training corpus. However, this is naturally limiting because the training corpus itself is bounded in time to the point of its collection. 
As a result, recent work has studied how to adapt such models to new data without an expensive retraining phase. Methods range from using a semi-parametric methods with access to external memory~(e.g., \citealt{guu-etal-2020-realm}; \citealt{lewis-etal-2020-retrieval}), to continual learning~(e.g., \citealt{timelm}; \citealt{lazaridou-etal-2021-mind}) to parameter efficient fine-tuning~(e.g., \citealt{ben-zaken-etal-2022-bitfit}; \citealt{pfeiffer-etal-2021-adapterfusion}).

%Embedded world factual knowledge~\citep{timelm} provides a natural demonstration for the diachronic degradation effect but only captures a fraction of the problem of LLMs becoming stale. 
%
Much of this work concerns factual knowledge or task distribution shifts. However, language also changes subtly: for instance, the popularity or meaning of individual words can change over time. In fact, such shifts also cause a consistent decrease in models performance for downstream tasks ~\citep{huang-paul-2018-examining,jaidka-etal-2018-diachronic,lukes-sogaard-2018-sentiment,florio-etal-2020}.

Acquiring new words through either examples or definitions is therefore an important test of LLMs' ability to overcome diachronic degradation. With in-context learning having emerged as the primary way to interact with LLMs~\citep{brown2020language}, we propose to study LLMs capability of acquiring new vocabulary via prompting. 
% for both language understanding and language generation tasks.

We propose \dataset{}, a novel benchmark for word acquisition for LLMs.
Word acquisition is challenging to study in a realistic setting as it is hard to know which terms a model has already been exposed to. To overcome this, we rely on a heuristic method to introduce newly invented words and define them in terms of existing concepts.
Following previous work~\cite{chakrabarty-etal-2022-rocket}, we incorporate the required knowledge into the prompt. 
We then ask models to perform tasks that require successfully interpreting the invented words.

\begin{figure*}[!th]
\centering
\begin{tabular}{p{7.1cm}p{0.05cm}p{7.1cm}}
\toprule
\winograd{} & & \\
\midrule 
The \textcolor{blue}{\underline{city councilmen}} refused the \textcolor{red}{demonstrators} a permit 
because \emph{they} \textbf{feared} violence. & & The \textcolor{red}{city councilmen} refused the \textcolor{blue}{\underline{demonstrators}} a permit 
because \emph{they} \textbf{advocated} violence. \\
\midrule
\dataset{} & \\
\midrule
The verb to \textbf{plest} means to be scared of, or want to avoid an object. & & The verb to \textbf{sparn} means to to publicly recommend or support. \\
The \textcolor{blue}{\underline{city councilmen}} refused the \textcolor{red}{demonstrators} a permit because \emph{they} \textbf{plested} violence. & & The \textcolor{red}{city councilmen} refused the \textcolor{blue}{\underline{demonstrators}} a permit because \emph{they} \textbf{sparned} violence. \\
\end{tabular}
\caption{An example pair from \dataset{} together with its original \winograd{} source. The task is to decide whether \emph{they} refers to the city councilman or the demonstrators. Here, the correct answer is shown in \textcolor{blue}{\underline{blue}} and the incorrect answer in \textcolor{red}{red}. Note that in both cases, it is necessary to understand the meaning of the bolded key concept to resolve the co-reference, which we identify in \winograd{} and substitute for a new word in \dataset{}.}
\label{fig:example}
\end{figure*}

% \begin{figure*}[!t]
% \centering
% \begin{tabular}{lp{10cm}}
% \winograd{}: & The city councilmen refused the demonstrators a permit 
% because \emph{they} \textbf{[feared | advocated]} violence. \\
% \dataset{}: & The verb to \textbf{plest} means to be scared of, or want to avoid an object. \\
%             & The verb to \textbf{sparn} means to to publicly recommend or support. \\
%             & The city councilmen refused the demonstrators a permit because \emph{they} \textbf{[plested | sparned]} violence. \\
% \end{tabular}
% \caption{An example pair from \dataset{} together with its original \winograd{} source. The meaning of the new concepts is critical to correctly resolve the co-reference in each case from the pair. The task is to decide whether \emph{they} refers to the city councilman or the demonstrators in each case.}
% \label{fig:example}
% \end{figure*}

We consider the co-reference resolution datasets Winograd Schema Challenge~\citep{winograd} and WinoGrande~\citep{winogrande}.
The examples are built in pairs with minimal changes, which allow the identification of the key concept that must be understood to solve the example. An example of \dataset{} can be seen in Figure~\ref{fig:example}.
Our contributions are the following:

\noindent \textbf{(a)} We propose \dataset{}, a method and dataset to test models for word acquisition skills.

\noindent \textbf{(b)} We benchmark the performance of several state-of-the-art models across scale and number of shots.

\noindent \textbf{(c)} We analyze the effect of prompt, POS tags, word likelihood and similarity for ease of acquisition.

These results help us understand the challenges for incorporating new concepts into LLMs. 
The code to build the dataset will be open-sourced
\footnote{\href{https://github.com/google-research/language/tree/master/language/wino_dict}{https://github.com/google-research/language/tree/master/language/wino\_dict}}.

\section{Methods}\label{sec:methods}

\dataset{}, like \winograd{} and \winogrande{}, is a co-reference resolution task in a binary choice setup. A model is given two alternative noun phrases, and has to decide which one is more likely to correspond to a highlighted pronoun or blank.

\subsection{Dataset Construction}
To build \dataset{}, we rely on the fact that the examples from \winograd{} and \winogrande{} are constructed from contrasting pairs~\citep{gardner-etal-2020-evaluating, Kaushik2020Learning}. 
Each instance differs in a minimal way from its counterpart with the true label reversed.
This allows the identification of the key concept that needs to be parsed in order to resolve the task.
In \autoref{fig:example} for instance, the verbs \emph{fear} and \emph{advocate} correspond to the key concepts.

The differences between the two datasets are that \winogrande{} is larger, uses blanks instead of pronouns and the dataset has been filtered for  co-occurrence bias between the key concept and the correct noun-phrase which unpairs some examples.

To create our examples, we first recover the pairing between the examples, dropping those with no pairing.
%\footnote{Note that the \winogrande{} debiasing process unpairs some examples, which are dropped in our process.}
%
Secondly, we identify the key concept tokens that change from one example to the other, dropping examples where the key concept consists of multiple tokens.
Finally we run the sentence through the spaCy\footnote{\href{https://spacy.io}{https://spacy.io}} syntactic analyzer and fetch WordNet\footnote{\href{https://wordnet.princeton.edu}{https://wordnet.princeton.edu}~\citep{Miller1995Wordnet}} definitions of the key concepts' lemmas.
In the next section we show how the key concept tokens are replaced by synthetic words.
This results in 496 examples: additional information can be found in \autoref{table:stats}.

\begin{table} % [!h]
\footnotesize
\begin{tabular}{l|rrr}
\toprule
\textbf{POS} & \textbf{\winograd{}} & \textbf{\winogrande{}} & \textbf{Total}\\
\midrule
VERB & 67	& 56	& 123 \\
NOUN & 34	& 24	& 58  \\
ADV  & 5	& 25	& 30  \\
ADJ  & 74	& 211	& 285 \\
\textbf{Total} & 180	& 316	& \textbf{496} \\
\midrule
\textbf{Orig. Size} & 273	& 12,282	& 12,555 \\
\textbf{Sent. Len} & 16.34 & 18.93 & 17.99 \\ 
\textbf{Def. Len} & 14.07 & 14.3 & 14.22 \\ 
\bottomrule
\end{tabular}
\caption{Statistics for the different part-of-speech tags in the synthetic words, as well as average number of tokens for the main statement and the word definition. \dataset{} consists of 496 examples.}
\label{table:stats}
\end{table}

\subsection{New word creation}

Our goal is to create plausible synthetic words. 
We create plausible words using a simple probabilistic model of every one-, two-, and three-letter sequences that was trained on the vocabulary of English words\footnote{\href{https://pypi.org/project/english-words}{https://pypi.org/project/english-words}}. These three-letter sequences are then sampled and combined to form new synthetic words.
We filter any words that have a three letter sequence that does not occur in any other English word. 
We then sample the words based on their log probability, placing them into five buckets and keeping around 500 for each bucket. 

The morphology for each word is created by aggregating over a sample of proposed synthetic word morphologies. The last 2-4 letters of each word (depending on the morphological edit) form a suffix dictionary that is used as a simple substitution dictionary for the remaining words: failures are dropped. This produces a combination of regular and irregular conjugations over the new words. 

\subsection{Answer Scoring}

Each instance in \dataset{} consists of a new word with its definition $d$, a statement containing a blank where $x$ and $y$ correspond to the text before and after the blank respectively, and two noun phrases $o_1$ and $o_2$. 
The task consists of identifying which of the noun-phrases better fits the blank.

\palm{}, \gptt{} and its predecessors~\citep{radford2019language} use the method proposed by~\citet{Trinh2018Simple} to evaluate \winograd{} and \winogrande{}, which we explain below.
A prediction score is obtained comparing the log likelihood of the same continuation $y$ of two possible prefix texts ($x:o_1$ and $x:o_2$) where the co-reference pronoun or blank marker has been replaced. 
% An example consists of two possible prefix texts, each representing one interpretation of the co-reference. 
%
% The model scores the same suffix text $y$ for each prefix: 
It is correct if it scores the suffix higher for the prefix with the correct interpretation of the co-reference problem.

\begin{tcolorbox}
\vspace{-1.0em}
\footnotesize
{
% \[\log P_\Theta\left(y|x:o_1\right) - \log P_\Theta\left(y|x:o_2\right)\]
% }
\begin{align*}
 \ln P_\Theta\left(y|x:o_1\right) &- \ln P_\Theta\left(y|x:o_2\right) \\
 \hspace{-0.5em}\sum_{i=0}^n \ln P_\Theta\left(y_i|y_{<i}:x:o_1\right) &- \ln P_\Theta\left(y_i|y_{<i}:x:o_2\right)
\end{align*}
\hspace{-0.5em}
}
where $:$ denotes concatenation and variables map to:
\begin{align*}
x =~ &\text{``The city councilmen refused the } \\
    &\text{  ~demonstrators a permit because''} \\
o_1 =~ &\text{``the city councilmen''} \\
o_2 =~ &\text{``the demonstrators''} \\
\{y_i\}_{i=1}^n = y =~   &\text{``feared violence.''}
\end{align*}
\end{tcolorbox}

\noindent In our setup we add the definition of the new concept as a suffix to the shared term $y$, thus replacing it with $y:d$ as this works the best. Note that this means that the model is \textit{scoring} the definition rather than conditioning on it.

See Section~\ref{sec:prompts} and \autoref{tab:baselines} for a discussion of other variants of the setup, including adding the definition as a prefix.

%

% \begin{equation}
%     \sum_{i}
% \end{equation}
\begin{table*}[!th]
\footnotesize
\centering
\resizebox{\textwidth}{!}{
\begin{tabular}{l||rrr|rrr||rrr|rrr}
%  & \multicolumn{6}{c}{\dataset{}} & \multicolumn{6}{c}{Original} \\
%  & \multicolumn{3}{c}{\winograd{}} & \multicolumn{3}{c}{\winogrande{}} &
%  \multicolumn{3}{c}{\winograd{}} & \multicolumn{3}{c}{\winogrande{}} \\
 & \multicolumn{6}{c}{\winograd{}} & \multicolumn{6}{c}{\winogrande{}} \\
 & \multicolumn{3}{c}{\dataset{} (Ours)} & \multicolumn{3}{c}{Original} &
 \multicolumn{3}{c}{\dataset{} (Ours)} & \multicolumn{3}{c}{Original} \\
\toprule
Shots & 0 & 1 & 5 & 0 & 1 & 5 & 0 & 1 & 5 & 0 & 1 & 5 \\
\midrule
\palm~8B & 59.2\err{1.6} & 57.1\err{2.1} & 59.1\err{1.6} & 83.3 & 83.3 & 87.2 & 51.8\err{1.6} & 54.2\err{0.4} & 52.4\err{1.1} & 69.3 & 65.5 & 67.4 \\
\palm~62B & 62.2\err{0.6} & 65.9\err{3.6} & 70.3\err{1.3} & 91.1 & 90.0 & 92.2 & 56.7\err{1.1} & 58.2\err{1.0} & 59.7\err{1.1} & 76.6 & 77.8 & 78.2 \\
\palm~540B & \textbf{65.9}\err{2.5} & \textbf{75.4}\err{1.3} & \textbf{78.6}\err{0.6} & 92.8 & 92.2 & 95.6 & \textbf{60.3}\err{1.4} & \textbf{63.9}\err{2.3} & \textbf{68.5}\err{1.9} & 80.1 & 81.3 & 85.8 \\
\midrule
\gptt~Ada & 51.9\err{2.2} & 50.9\err{1.7} & 50.2\err{4.3} & 60.0 & 57.8 & 61.7 & 52.2\err{1.2} & 52.0\err{3.6} & 49.4\err{1.7} & 48.1 & 53.8 & 53.2 \\
\gptt~Babbage & 51.8\err{0.8} & 52.8\err{2.0} & 54.4\err{2.3} & 75.6 & 71.7 & 65.6 & 50.8\err{1.7} & 52.3\err{1.0} & 52.2\err{0.8} & 52.8 & 55.1 & 56.6 \\
\gptt~Curie & 54.2\err{1.6} & 54.6\err{2.4} & 59.9\err{1.5} & 85.0 & 81.7 & 82.8 & 50.2\err{1.5} & 50.6\err{1.6} & 52.2\err{1.0} & 62.0 & 61.1 & 60.8 \\
\gptt~Davinci & 60.3\err{1.3} & 63.6\err{2.3} & 72.9\err{0.5} & 88.3 & 85.0 & 91.1 & 55.0\err{1.1} & 55.7\err{1.4} & 61.3\err{1.4} & 71.8 & 69.6 & 72.5 \\
\midrule
Human & 91.7 &  &  & 96.5$^*$ &  &  & 83.3 &  & & 94.0$^*$ & \\
\bottomrule
\end{tabular}}
\caption{Binary classification accuracy on \dataset{} vs. the original datasets using average and standard deviation across $5$ sets of new words. Original results may differ from the ones reported by~\citet{chowdhery2022palm} since only a subset of the examples are used. A consistent gap of $18+$ points appears when comparing against the original sets.
The original human evaluation numbers denoted with $^*$ are taken from~\citet{winogrande}.}
\label{table:results}
\end{table*}

\section{Experiments}
In this work we test 
% T5 \citep{raffel-etal-2020-exploring},
\gptt~\citep{brown2020language}, and \palm~\citep{chowdhery2022palm} models of various sizes, ranging from 3B to 540B parameters. \autoref{app:model-sizes} has more details on the model sizes.

As in the original in-context learning evaluations, we try $0$, $1$, and $5$-shot experiments, using random examples to build the prompt. We compare to both a zero-shot human evaluation as well as a the original setting with only our filtered examples.

The main experimental results are shown in \autoref{table:results}. We observe a consistent gap of 18 or more points between \dataset{} examples and their original counterparts.
Similar to trends observed in other datasets~\citep{chowdhery2022palm}, scaling the number of shots and model size consistently improves accuracy. 
%
%\gptt{}-Davinci performs comparably to \palm{}-62B but generally manages better results on the $5$-shot case. 
The three smaller versions of \gptt{} and \palm{}-8B all perform close to random.

% We also experimented with other prompting strategies and baselines.
%
We verify that omitting any information of the new word yields random results for even the best \palm{}-540B model. We discuss this and other prompting strategies in more detail in \autoref{app:prompts}.

\subsection{Human Evaluation}

The human accuracy on \dataset{} is estimated using the responses of $10$ volunteers.
No native English proficiency was required for participation.
Participants were told that the aim of the research is to study how to use words based on their definition.
They were presented with $15$ sentences that included a pronoun / blank and asked to select one of two noun phrases it most likely refers to. 
\begin{table}[th!]
\footnotesize
\centering
\resizebox{0.45\textwidth}{!}{
\begin{tabular}{llrr}
\toprule
Word Type & Prompt &   \winograd{} &  \winogrande{} \\
\midrule
Synthetic & Def Prefix &  72.2 &   62.7 \\
      & Def Suffix$^*$ &  78.6 &   68.5 \\
      & Syn Prefix &  74.1 &   60.5 \\
      & Syn Suffix &  88.4 &   78.2 \\
      & Empty &  52.0 &   51.9 \\
\midrule      
Original & Def Prefix &  85.5 &   74.0 \\
      & Def Suffix &  93.8 &   84.4 \\
      & Syn Prefix &  87.2 &   74.3 \\
      & Syn Suffix &  91.6 &   83.2 \\
      & Empty$^*$ &  95.6 &   85.8 \\
\midrule      
Meaning shift & Def Prefix &  66.1 &   60.8 \\
          & Def Suffix &  75.6 &   60.4 \\
          & Syn Prefix &  69.4 &   60.1 \\
          & Syn Suffix &  83.3 &   74.7 \\
          & Empty &  51.1 &   49.7 \\
\bottomrule
\end{tabular}}
\caption{Analysis of different prompts. We show the results on the synthetic words, original words, and existing words but assigned to a new meaning (``Meaning shift''). Prefix/Suffix correspond to the location of the definition, Syn/Def corresponds to using the definition or synonyms of the synthetic word. Empty means neither (should be random for synthetic words). Providing synonyms yields the best results. All results are on \palm{}-540B $5$-shot. The lines marked with $^*$ correspond to the experiments in \autoref{table:results}.}
\label{tab:baselines}
\end{table}
\section{Prompt Analysis}\label{sec:prompts}

In this section, we discuss alternative formulations for the prompts used in \dataset{}. We focus on the best-performing \palm-{540B} model using a 5-shot setup. See \autoref{tab:baselines} for the full results.

Concretely, we vary the prompts along a few axes. First, we test whether the definition should be part of the prefix, where the model would condition on it, or the suffix, where the model would score it.
Note that in all setups, putting the definition in the suffix works consistently better.

Additionally, we test whether the task is made easier by using synonyms instead of definitions. This task indeed appears to be easier, potentially because the model needs to learn only a simple substitution between the new word and old word. We focus on definitions in this work as exact synonyms would rarely be available for novel words.

As a baseline, we also examine the ``Empty'' setup, where the model is provided no information about the new word. We observe that \palm{} approximates random guessing without being given the definition, showing that the task remains roughly unbiased. 

We additionally test the model's performance on the original task where the definition is provided. Note that the ``Empty'' case here corresponds precisely to the original task. Interestingly, the definition seems to serve as a slight distraction, especially as a prefix, though accuracy is still well above the model's performance on the synthetic words.

Finally, in the ``Meaning shift'' scenario, we map new definitions to already known words. This task appears to be even more difficult than the standard \dataset{} setup, implying that the model is distracted by the surface forms of the words.

\section{New Word Analysis}

Several factors can affect the capabilities for word acquisition of LLMs.
%
%In order to evaluate this effect, we compiled a list of attributes that could be predictive of accuracy, split the resulting data in quartiles according to each attribute and evaluated the model results.
%
We investigate several attributes, split into quartiles, using \palm{}-540B with $5$-shots, which is the best model from \autoref{table:results}.

We consider the following attributes. 

\begin{enumerate}
\item{The part-of-speech of the synthetic word.}
\item{The average model negative log likelihood (NLL) of the two model predictions, which measures the likelihood of the suffix for both prefixes.} 
\item{The number of SentencePiece~\citep{kudo-richardson-2018-sentencepiece} tokens in the synthetic word, to investigate the effect of model tokenization.}
\item{The number of SentencePiece tokens in the definition of the synthetic word, to investigate if longer definitions are more challenging.}
\item{The Levenshtein edit distance between the synthetic and original word, to investigate if similar words are easier.} 
\item{The likelihood of the new word as computed by our probabilistic model of three-letter sequences, to see if less probable words are more difficult to acquire.}
\end{enumerate}

\begin{figure}[ht]
    \centering
    \includegraphics[width=0.95\linewidth]{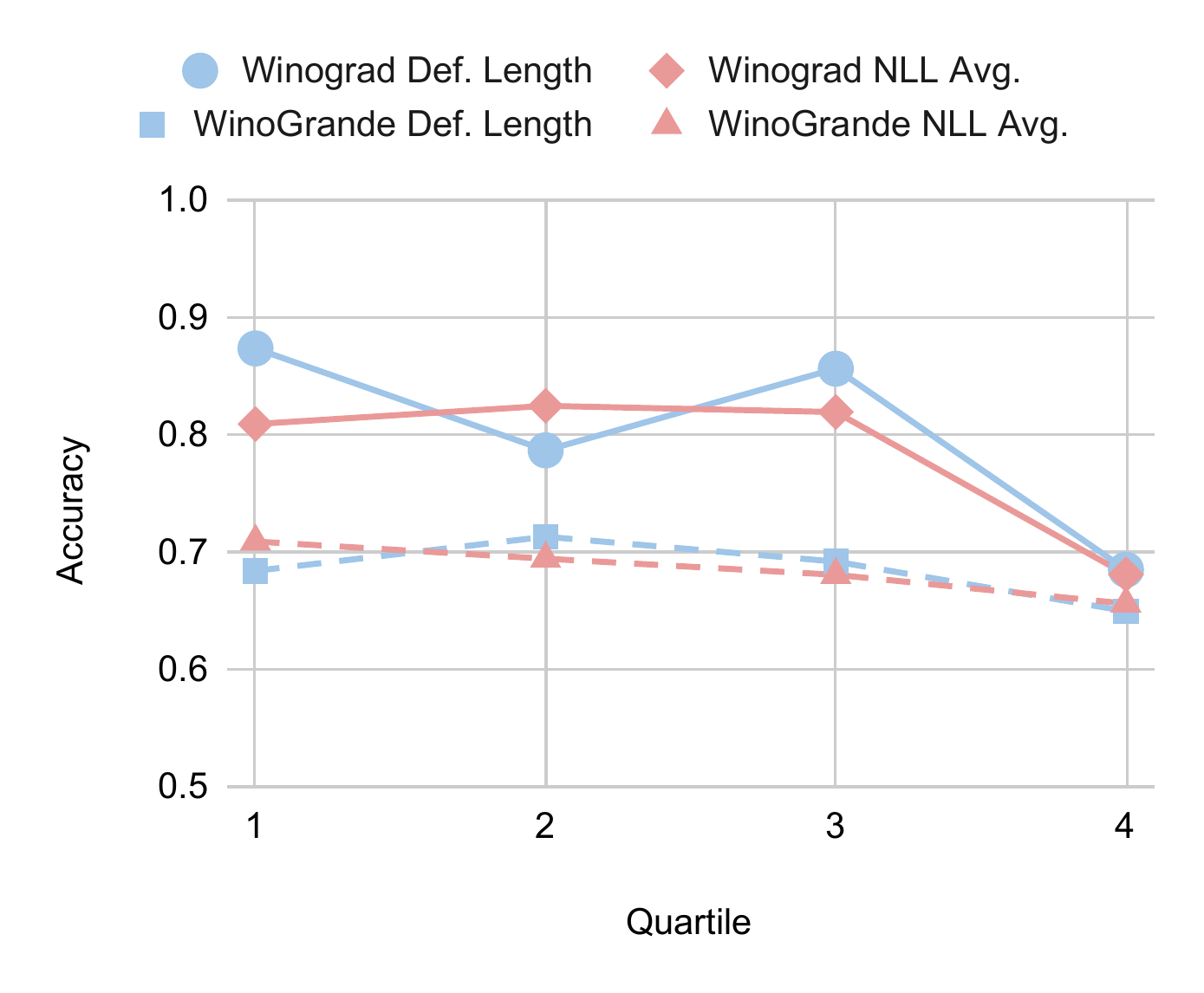}
    \caption{Effect on \dataset{} \palm{}-540B $5$-shot accuracy on each quartile splitting by definition length and by average NLL score. Longer definitions and higher NLL correlate with lower accuracy.}
    \label{fig:breakdown}
\end{figure}

Of the six attributes, the two most correlated with accuracy are (4) the definition length and (2) the average NLL. We observe no clear pattern in the other four attributes.
In Figure~\ref{fig:breakdown} we show their effect in each quartile.
The effect of definition length indicated that the $25\%$ longest definitions are the hardest to acquire by a significant margin ($12\%$ relative drop for \winograd{}, $5\%$ for \winogrande{}).
The relative accuracy drop for the largest quartile of the NLL average is $13\%$ for \winograd{} and $4\%$ for \winogrande{}. The drop in NLL suggests that when models assigns low probabilities to answers, they make more mistakes: the low probability may indicate the model has a poor understanding of the prefix so scores the suffix randomly.

%which may be hurting its discriminative power due to uncalibrated predictions far from the mode.

\section{Limitations}

The task described in this work is synthetic and thus an imperfect measure of the phenomena under study.
The words in \dataset{} are synthetic words with definitions copied from existing concepts; the model could thus solve \dataset{} with a reduction to a reverse dictionary task. To partially address this, we conduct a pilot experiment using twenty hand-written \dataset{}-like examples whose definitions are instead inspired by foreign words that do not have a clear single-word definition in English. For instance, ``estrenar'' refers to wearing a piece of clothing for the first time, which does not have a clear English word equivalent. We can then create an example that requires knowing this definition, such as ``I really [ \underline{love} | hate] my new dress. I can’t wait to <word> it.

\begin{table}[!th]
\footnotesize
\centering
\begin{tabular}{l||rrr}
\toprule
Shots & 0 & 1 & 5 \\
\midrule
\palm{}~540B & \textbf{68.7} & \textbf{73.0} & \textbf{76.0}  \\
\gptt~Davinci & 61.0 & 56.0 & 68.0 \\
\bottomrule
\end{tabular}
\caption{Binary classification accuracy on the foreign-inspired new words averaged over five runs. Overall, accuracy is comparable to the original dataset.}
\label{tab:foreign_results}
\end{table}

In conducting this experiment, we substitute synthetic words instead of using the original foreign words, and the definitions of the words themselves may not correspond to native speakers' precise understanding: in other words, these are 
%still 
intended to be genuine new words and data leakage should be minimal. We run the same experiment on these examples. Results are in \autoref{tab:foreign_results} and full details are in \autoref{app:foreign}. Overall, the numbers are comparable to the original dataset, suggesting that models are unlikely to be solving the problem in this way. 

% Additionally, the definition may contain additional clues, such as the typical subject or object of a verb; however, the de-biasing process performed by~\citet{winogrande} in \winogrande{} would have most likely discarded such occurrences.
%
% Nonetheless, the large gap in performance between the real and synthetic shows the value of the benchmark in identifying future challenges for LLMs.

Finally, the choice of prompts for LLMs has been shown to have a large influence on the resulting accuracy \citep{min-etal-2022-rethinking, lu-etal-2022-fantastically}. 
While we tried multiple templates it is possible that substantially better prompts exist for this task.
\section{Related Work}

\paragraph{Word acquisition for LLMs.}
Inspired by developmental linguistics ~\citep{carey1079Acquiring}, \citet{radford2019language} succeeded to prompt \gptt{} to generate plausible example sentences based on definitions of synthetic words. Unlike \dataset{}, the evaluation was purely qualitative.

% \noindent\textbf{Knowledge Enhanced Prompts}
% \citet{mishra-etal-2022-cross}

\paragraph{Common sense.}
\citet{li-etal-2021-systematic} study how prompt structures and scoring methods affect the performance of LLMs on common sense tasks including \winogrande{}, where they observe the least variation.
%
%\citet{chakrabarty-etal-2022-rocket} study story continuations using figurative language and observe benefits in injecting common sense information from COMET-ConceptNet~\citep{Hwang2021Comet}.
The format from \winograd{} has been subsequently used to probe models for other phenomena such as explanations~\citep{zhang-etal-2020-winowhy} and gender bias~\citep{zhao-etal-2018-gender}.

\paragraph{Benchmarks for lexical knowledge.}
\citet{rare_words} introduce a benchmark for probing a model's knowledge of the properties of rare words.
\citet{hill-etal-2016-learning-understand} train models to match word and definition representations, which they apply to a reverse dictionary task.
\section{Conclusion}

In this work, we study the question of in-context word acquisition by large language models. 
While non-trivial to measure, the ability to incorporate knowledge about new words in-context may be useful to decrease the effect of diachronic degradation.
% or improve the performance in multilingual tasks.
%
% In order to measure progress on this relevant but highly subjective task, 
We design a mechanism to transform Winograd-style tasks into challenging probes for reasoning on the meaning assigned to synthetic words, allowing for a more objective measurement of word acquisition.
We study the results of models of multiple sizes and families and conclude that while the problems becomes easier with scale, there is still a substantial gap with human performance and the original \winograd{} and \winogrande{} tasks, demonstrating the difficulty of the proposed task.
Finally, we show that acquiring novel definitions is of similar difficulty, indicating the task is realistic.
\section*{Acknowledgements}

We thank Yasemin Altun, Iulia-Maria Comşa and Srini Narayanan, as well as our anonymous reviewers, for their valuable feedback.

\bibliography{anthology,custom}
\bibliographystyle{acl_natbib}

\clearpage
\appendix

\section{Model Sizes}\label{app:model-sizes}
While OpenAI does not officially disclose the size of their four models Davinci, Curie, Babbage and Ada, we use the numbers approximated in a blogpost as estimates.\footnote{\href{https://blog.eleuther.ai/gpt3-model-sizes}{https://blog.eleuther.ai/gpt3-model-sizes}} 
\autoref{tab:model_size} contains the number of parameters for the models used in our experiments.

\begin{table}[!ht]
    \centering
    \scalebox{0.78}{
    \begin{tabular}{lc}
    \toprule
    Model & \# Parameters\\
    \midrule
    \gptt{}-Ada & 350M  \\
    \gptt{}-Babbage & 1.3B  \\
    \gptt{}-Curie & 6.7B  \\
    \gptt{}-Davinci & 175B \\
    \midrule 
    \palm{}-8B & 8B \\
    \palm{}-62B & 62B \\
    \palm{}-540B & 540B \\
    \bottomrule
    \end{tabular}
    }
    \caption{Number of parameters of the reported models.}
    \label{tab:model_size}
\end{table}

\section{Prompts}\label{app:prompts}
We built prompts for definitions and synonyms to make them sound natural given the structure of most WordNet definitions for each part-of-speech tag.
Table~\ref{tab:prompts} shows the different prompt templates in each case.

\begin{table}[!ht]
    \centering
    \footnotesize
    % \scalebox{0.78}{
    \begin{tabular}{lp{3.5cm}}
    \toprule
    Type & Prompt\\
    \midrule
    Synonym & The meaning of \{lemma\} is similar to \{synonym\}. \\
    Verb definition & The verb to \{lemma\} means to \{definition\}. \\
    Noun definition & The word \{lemma\} refers to \{definition\}. \\
    Adj. Definition & The meaning of \{lemma\} is {definition}. \\
    Adv. Definition & The word \{lemma\} means \{definition\}. \\
    \bottomrule
    \end{tabular}
    % }
    \caption{Number of parameters of the reported models.}
    \label{tab:prompts}
\end{table}

% \section{Dataset Statistics}\label{app:data-stats}
% A part of speech breakdown for each dataset can be found in \autoref{table:stats}.

% \begin{table} % [!h]
% \footnotesize
% \begin{tabular}{l|rrr}
% \toprule
% \textbf{POS} & \textbf{\winograd{}} & \textbf{\winogrande{}} & \textbf{Total}\\
% \midrule
% VERB & 67	& 56	& 123 \\
% NOUN & 34	& 24	& 58  \\
% ADV  & 5	& 25	& 30  \\
% ADJ  & 74	& 211	& 285 \\
% \textbf{Total} & 180	& 316	& \textbf{496} \\
% \midrule
% \textbf{Orig. Size} & 273	& 12,282	& 12,555 \\
% \textbf{Sent. Len} & 16.34 & 18.93 & 17.99 \\ 
% \textbf{Def. Len} & 14.07 & 14.3 & 14.22 \\ 
% \bottomrule
% \end{tabular}
% \caption{Statistics for the different part-of-speech tags in the synthetic words, as well as average number of tokens for the main statement and the word definition. \dataset{} consists of 496 examples.}
% \label{table:stats}
% \end{table}

\section{Foreign Inspired Words}\label{app:foreign}
In \autoref{tab:foreign_words} we list the word, approximate definition, and \dataset{}-like example. Note that these examples are handwritten and did not go through a debiasing process like \winogrande{} In order to reduce the risk of data leakage, in the actual examples we replace the surface form of the word with one of the synthetic surface forms using the same process as in \autoref{sec:methods}. 

% \section{Other Prompting Strategies}

\begin{table*}[th]
\footnotesize
\centering
\begin{tabular}{p{9cm}|p{6cm}}
Example & Definition \\
\toprule
John frequently goes backpacking and Jake never does because [ \underline{Jake} | John] disdains the feeling of \textbf{waldeinsamkeit}. & the feeling of solitude and connectedness to nature when being alone in the woods \\
\midrule
After returning from backpacking, John thought he would go again [ \underline{frequently} | infrequently]. John likely appreciates the feeling of \textbf{waldeinsamkeit}. & the feeling of solitude and connectedness to nature when being alone in the woods \\
\midrule

Mary loves going to antique stores and Ashley never does because [ \underline{Mary} | Ashley] \textbf{wabi-sabis} old things. & finding beauty in imperfections \\
\midrule

Mary loves going to [ \underline{antique} | modern] stores because she \textbf{wabi-sabis} old things. & finding beauty in imperfections \\
\midrule

Pierre is from France and John is from Ireland. Pierre and John like to go to Irish bars and talk about [ \underline{Pierre} | John]'s feeling of \textbf{depaysement} there. & the feeling that comes from not being in one's home country; being a foreigner \\
\midrule

Pierre has lived in France all his life. When he's in [ \underline{Ireland} | France], Pierre frequently talks about his feeling of \textbf{depaysement}. & the feeling that comes from not being in one's home country; being a foreigner \\
\midrule

Jake and Ashley plan to get married, Ashley's parents are happy, but Jake's parents don't like it because a friend said they had bad \textbf{yuanfen}. [ \underline{Jake} | Ashley]'s parents are more likely to go to a fortune teller. & the fate between two people \\
\midrule

Jake and Ashley plan to get married. Ashley's parents are very practical while Jake's parents believe in destiny. When an advisor said Jake and Ashley had bad \textbf{yuanfen}, [ \underline{Jake} | Ashley] wanted to call it off. & the fate between two people \\
\midrule

Theresa doesn't get why Martha thinks the statue in the museum was so \textbf{duende} that [ \underline{Martha} | Theresa] spent a lot of time looking at it. & a work of art's mysterious power to deeply move a person \\
\midrule

Martha spends a lot of times in museums while Theresa spends little. [ \underline{Martha} | Theresa] finds art \textbf{duende}. & a work of art's mysterious power to deeply move a person \\
\midrule

After losing his [ \underline{religion} | job], John fell into a sense of \textbf{toska}. & a sensation of great spiritual anguish, often without a specific cause; a longing with nothing to long for \\
\midrule

John kept yelling at Joey for not doing chores, but Joey wouldn't even respond. [ \underline{Joey} | John] really seems \textbf{tosked}. & a sensation of great spiritual anguish, often without a specific cause; a longing with nothing to long for \\
\midrule

Because he [ \underline{loves} | hates] reptiles, John found seeing that group of lizards very \textbf{gigil}. & a situation of such extreme cuteness it's overwhelming or the irresistable urge to hug something cute \\
\midrule

John only keeps salamanders as pets and Joey likes more traditional ones, so [ \underline{John} | Joey] found seeing the group of lizards very \textbf{gigil}. & a situation of such extreme cuteness it's overwhelming or the irresistable urge to hug something cute \\
\midrule

John thought his marriage with Joey was \textbf{shougani}, so he wanted to hire a [ \underline{lawyer} | therapist]. & a situation that can't be helped, or an act of resignation \\
\midrule

John thought his marriage with Joey was \textbf{shougani} but Joey disagreed, so [ \underline{John} | Joey] decided to hire a lawyer. & a situation that can't be helped, or an act of resignation \\
\midrule

Joey still can't get over when John drunkenly called him Mark at his wedding, and now whenever they see each other, [ \underline{Joey} | John] \textbf{tartles}. & a moment of hesitation when introducing someone because you can't remember their name \\
\midrule

I really [ \underline{love} | hate] my new dress. I can't wait to \textbf{estrene} it. & wearing something for the very first time \\
\midrule

Mary and Sue went dress shopping together. Mary hates her dress while Sue loves hers. [ \underline{Sue} | Mary] can't wait to \textbf{estrene} it. & wearing something for the very first time \\
\midrule

After a long day of work, James \textbf{xinkued} the job John did. John was [ \underline{grateful} | upset]. & acknowledging someone's effort for working hard or doing you a favor \\
\bottomrule
\end{tabular}
\caption{List of foreign-inspired new words (\textbf{bolded}) and their corresponding examples and definitions. The possible choices for the example are shown, with the correct choice underlined. The definition is shown on the right. These definitions may or may not be idiosyncratic to a native speaker; however, the actual examples use a synthetic word to more closely resemble new word acquisition and minimize the risk of data leakage.}
\label{tab:foreign_words}
\end{table*}

\end{document}